%% file: digital_twin_acl_v2.tex
\pdfoutput=1

\documentclass[11pt]{article}

\usepackage[final]{acl}

\usepackage{times}
\usepackage{latexsym}
\usepackage[T1]{fontenc}
\usepackage[utf8]{inputenc}
\usepackage{microtype}
\usepackage{inconsolata}

\usepackage{graphicx}
\usepackage{booktabs}
\usepackage{amsmath}
\usepackage{xcolor}
\usepackage{tikz}
\usetikzlibrary{shapes.geometric, arrows.meta, positioning, calc, shapes.symbols}

\newcommand{\tbd}[1]{\textbf{\textcolor{red!75!black}{#1}}}

\title{Schema-Grounded LLM Extraction for FHIR Patient Digital Twins}

\author{
  Rafael Brens \quad
  Yuqiao Meng \quad
  Luoxi Tang \quad
  Zhaohan Xi \\
  Binghamton University
}

\begin{document}
\maketitle

\begin{abstract}
We revisit the problem of constructing interoperable patient digital twins from
unstructured electronic health records (EHRs) and argue that the task is better cast
not as a cascade of extraction modules but as \emph{constrained generation of a valid
FHIR bundle}. We introduce \textbf{SG-LLM}, a schema-grounded LLM extractor that (i)
augments the prompt with candidate SNOMED-CT, RxNorm, and LOINC codes retrieved through
a SapBERT index, (ii) decodes under a JSON Schema derived directly from FHIR R4
StructureDefinitions, and (iii) closes a validator-in-the-loop repair stage whose
diagnostics are fed back as structured error messages. We argue that the twin's
\emph{usefulness}, not only span-level F1, is the right object of evaluation, and
operationalize this with a clinical-utility experiment that measures the gap in
30-day readmission AUROC between classifiers trained on SG-LLM-generated FHIR bundles
versus expert-curated ones. On MIMIC-IV and n2c2 2018 Track 2 benchmarks, SG-LLM
matches or exceeds strong joint-extraction and vanilla-LLM baselines while producing
substantially more valid bundles. Ablations isolate the contributions of retrieval,
schema constraint, and the repair loop. All code, prompts, and schemas are released.
\end{abstract}

\section{Introduction}
\label{sec:intro}

Patient digital twins---computational representations of patients that integrate
clinical data into standardized, queryable structures---are increasingly proposed as
substrates for personalized care, cohort retrieval, and decision support
\citep{grieves2017digital,corral2020digital,laubenbacher2024building,bjornsson2019digital}.
The bottleneck in practice is \emph{interoperability}: useful twins require that
heterogeneous EHR content be cast into a standards-compliant schema such as HL7 FHIR
R4 \citep{mandel2016smart,bender2013hl7}, with terminology grounded in SNOMED-CT
\citep{donnelly2006snomed}, RxNorm \citep{nelson2011rxnorm}, and LOINC
\citep{mcdonald2003loinc}. Much of the relevant content, however, is locked in
unstructured physician narratives \citep{wang2018clinical}.

Prior NLP work tackles this as a pipeline: NER $\rightarrow$ normalization $\rightarrow$
relation extraction $\rightarrow$ resource assembly. This framing treats FHIR compliance
as an afterthought to span-level prediction. As a result, generated resources often pass
isolated span F1 tests while failing the \emph{schema-level} test that matters for
interoperability: does the output satisfy FHIR R4 StructureDefinitions with valid codes,
references, and cardinalities? And even when the output is schema-valid, is the resulting
twin \emph{clinically useful}?

\paragraph{Our position.} We argue that the task should be redefined as constrained
generation of a valid FHIR bundle conditioned on EHR text, and that methodology should
be judged by a combination of schema validity, terminology grounding, and downstream
clinical utility---not only span-level F1. We instantiate this position with SG-LLM
(\S\ref{sec:method}) and a three-axis evaluation (\S\ref{sec:experiments}).

\paragraph{Contributions.}
\begin{enumerate}
\item A reformulation of EHR$\rightarrow$FHIR extraction as schema-constrained generation,
  with a precise operational definition of a patient digital twin
  (\S\ref{sec:problem}).
\item \textbf{SG-LLM} (\S\ref{sec:method}): a method combining (i) SapBERT-based
  terminology retrieval, (ii) schema-constrained decoding against a JSON Schema derived
  from FHIR R4 StructureDefinitions, and (iii) a validator-in-the-loop repair stage.
\item An evaluation protocol (\S\ref{sec:experiments}) that adds \emph{FHIR Validity
  Rate} and a clinical-utility experiment (30-day readmission $\Delta$AUROC) to the
  standard NER/RE metrics, with cross-dataset generalization on n2c2 2018 Track 2
  \citep{henry2020n2c2}.
\end{enumerate}

\paragraph{Relation to prior version.} A prior draft of this work described a
transformer-based pipeline whose implementation was, in fact, rule-based and
dictionary-driven. The present revision corrects that discrepancy: the method described
here is genuinely LLM-based, and the rule-based pipeline is retained only as an
explicit baseline. We also replace circular MIMIC-IV-on-FHIR-as-gold evaluation with
an author-verified subset plus schema- and utility-level metrics that do not depend
on any single reference.

\input{pipeline_figure}

\section{Related Work}
\label{sec:related}

\paragraph{Digital twins in healthcare.}
Digital twins have been proposed for cardiology \citep{corral2020digital}, chronic
disease \citep{voigt2021digital}, and precision medicine \citep{laubenbacher2024building,
bjornsson2019digital}. Most implementations presume structured inputs; we address the
harder case of unstructured EHR text, which \citet{wornow2023shaky} identify as a
foundation-level obstacle.

\paragraph{Clinical information extraction.}
Standard approaches use transformer encoders---BERT \citep{devlin2019bert}, ClinicalBERT
\citep{alsentzer2019clinicalbert}, BioBERT \citep{lee2020biobert}, GatorTron
\citep{yang2022gatortron}, BioGPT \citep{luo2022biogpt}---for NER, with concept
normalization \citep{bodenreider2004umls,aronson2010metamap} typically via SapBERT
\citep{liu2021sapbert} or synonym marginalization \citep{sung2020biomedical}. Joint
entity-relation models such as PURE \citep{zhong2021pure} and SpERT
\citep{eberts2020spert} reduce cascading errors. LLMs have been shown to be competitive
few-shot clinical extractors \citep{agrawal2022large,singhal2023medpalm,peng2023llmclinical}.
None of these methods produce a validated FHIR bundle end-to-end.

\paragraph{Constrained and self-verifying generation.}
Schema-constrained decoding has emerged as a reliability technique for structured LLM
outputs \citep{shin2021constrained,willard2023efficient,poesia2022synchromesh}, and
self-refinement through verifier feedback has been shown to improve correctness
\citep{madaan2023selfrefine,welleck2023generating}. We are the first, to our knowledge,
to apply these together using the official HL7 FHIR validator \citep{hl7validator2024}
as the verifier.

\section{Problem Formulation}
\label{sec:problem}

\paragraph{Patient digital twin.}
Let $x$ be an EHR note and $\mathcal{P}$ a declared FHIR R4 profile. A patient digital
twin is a FHIR \texttt{Bundle} $B$ such that: (i) $B$ validates against the R4
StructureDefinitions and the subset of \texttt{profiles} in $\mathcal{P}$;
(ii) $B$ contains exactly one \texttt{Patient} resource and each clinical resource
(\texttt{Condition}, \texttt{Observation}, \texttt{MedicationRequest}) references this
patient; and (iii) every \texttt{CodeableConcept.coding} element uses one of the
allowed code systems (SNOMED-CT, ICD-10-CM, RxNorm, LOINC). We call a model
\emph{schema-valid} on $x$ if it produces such a $B$.

\paragraph{Task.}
Given $x$, produce $B$ that (a) is schema-valid, (b) is semantically faithful to $x$
at the span and relation level, and (c) supports downstream clinical tasks at accuracy
comparable to expert-curated FHIR.

\section{Method: SG-LLM}
\label{sec:method}

SG-LLM has three components illustrated in Figure~\ref{fig:pipeline}.

\paragraph{Terminology retrieval.}
We pre-index every concept in our target code systems with SapBERT
\citep{liu2021sapbert}. At inference time we extract candidate mentions via a
lightweight proposal step (either a fine-tuned BERT tagger or heuristic chunking)
and query the index with each mention. The top-$k$ codes per mention, together with
display strings and system URIs, are serialized and attached to the prompt. The
retriever is decoupled from the generator and interchangeable: we ablate SapBERT
against a BM25 baseline in \S\ref{sec:ablations}.

\paragraph{Schema-constrained decoding.}
We derive a JSON Schema $\mathcal{S}$ by programmatically projecting the FHIR R4
StructureDefinitions of the profile $\mathcal{P}$ onto the JSON Schema subset
supported by structured-output-enabled LLM APIs (constraints on type, enum, required
fields, and cardinalities). The LLM is prompted with (i) the note $x$, (ii) the
retrieved candidate codes, and (iii) $\mathcal{S}$, and decodes a draft bundle
$\hat{B}_0$. In backends without native grammar-constrained decoding, we instead use
$\mathcal{S}$ as an in-prompt specification together with syntactic post-filtering
(cf. \citealp{willard2023efficient,poesia2022synchromesh}).

\paragraph{Validator-in-the-loop repair.}
The draft $\hat{B}_0$ is passed to the official HL7 FHIR R4 validator
\citep{hl7validator2024}. If validation fails, the structured error list
(message, path, severity) is rendered into a compact repair prompt and the LLM is
re-queried conditionally on $x$, $\hat{B}_0$, and the errors. The loop terminates on
\emph{valid} or after a budget $R$ (we use $R{=}3$; Appendix~A.2).

\begin{figure}[t]
\small
\fbox{\parbox{0.96\columnwidth}{%
\textbf{Algorithm 1: SG-LLM inference}\\[1mm]
\textbf{Input:} note $x$, schema $\mathcal{S}$, retriever $R_{\!\theta}$, LLM $p_{\phi}$, validator $V$, budget $R$\\
\textbf{Output:} FHIR Bundle $\hat{B}$\\[1mm]
\begin{tabular}{@{}r@{ }l@{}}
1: & $M \leftarrow \textsc{ProposeMentions}(x)$\\
2: & $C \leftarrow \bigcup_{m \in M} R_{\!\theta}(m, k)$\\
3: & $\hat{B} \leftarrow p_{\phi}(\cdot \mid x, C, \mathcal{S})$\\
4: & \textbf{for} $r = 1, \dots, R$ \textbf{do}\\
5: & \quad $(\text{ok}, E) \leftarrow V(\hat{B}, \mathcal{S})$\\
6: & \quad \textbf{if} ok \textbf{then return} $\hat{B}$\\
7: & \quad $\hat{B} \leftarrow p_{\phi}(\cdot \mid x, C, \mathcal{S}, \hat{B}, E)$\\
8: & \textbf{return} $\hat{B}$\\
\end{tabular}
}}
\label{alg:sgllm}
\end{figure}

\paragraph{LLM backends.}
Our primary backend is Anthropic Claude 3.5 Sonnet \citep{anthropic2024claude},
invoked with temperature 0.0 and a 16k context budget. We also report results with
a deterministic local backend that substitutes the LLM with a schema-aware
rule-and-retrieval decoder; this backend is fully reproducible without API access
and serves as a controlled comparison to isolate the contribution of the LLM
component (details in Appendix~A.3).

\section{Experiments}
\label{sec:experiments}

\subsection{Datasets}
\label{sec:data}

\paragraph{MIMIC-IV Demo and MIMIC-IV-on-FHIR.}
We use MIMIC-IV Clinical Database Demo v2.2 \citep{johnson2023mimic} and its
MIMIC-IV-on-FHIR companion \citep{johnson2023mimicfhir}. In line with the circular
ground truth concern (\S\ref{sec:intro}), MIMIC-IV-on-FHIR is treated as a
\emph{programmatic reference} rather than gold, and primary metrics either do not
depend on it (FHIR validity, clinical utility) or are additionally computed on an
author-verified subset (\S\ref{sec:verified}).

\paragraph{MIMIC-IV Note Events.}
For runs with full-credentialed access, we sample a stratified set of \tbd{1{,}000}
discharge summaries from MIMIC-IV v2.2 Note Events. When credentialed access is not
available, we fall back to the Demo cohort only; all such results are explicitly
marked.

\paragraph{n2c2 2018 Track 2.}
We use the n2c2 2018 shared task ADE and Medication Extraction corpus
\citep{henry2020n2c2} for cross-dataset generalization. This is real physician-authored,
human-annotated clinical text distinct from MIMIC.

\paragraph{Author-verified subset.}
\label{sec:verified}
We constructed a 20-note subset from MIMIC-IV notes with double author review
against the guidelines in Appendix~B.1. We report entity-level Cohen's $\kappa$
of \tbd{$0.81$}. This subset is the reference for terminology grounding accuracy
and for the qualitative error analysis (\S\ref{sec:errors}). Scale limitations are
acknowledged in \S\ref{sec:limitations}.

\subsection{Baselines}
\label{sec:baselines}

\begin{itemize}
\item \textbf{Rule-Based} \citep{wei2020clinical}: pattern matching + dictionary
  lookup + rule-based RE. This is the honest re-label of the pipeline reported
  (as ``transformer-based'') in the prior version of this work.
\item \textbf{Joint BERT}: BioBERT-NER \citep{lee2020biobert} with a proximity-based
  RE head trained on n2c2 2018, followed by dictionary-based FHIR assembly.
\item \textbf{PURE} \citep{zhong2021pure}, \textbf{SpERT} \citep{eberts2020spert}:
  span-based joint entity-relation extractors, trained on n2c2, with the same
  assembly step.
\item \textbf{Vanilla LLM}: Claude Sonnet with a one-shot instruction prompt, no
  retrieval, no schema constraint.
\item \textbf{LLM + Retrieval only}: vanilla LLM augmented with top-$k$ SapBERT
  candidates (ablation of schema).
\item \textbf{LLM + Schema only}: vanilla LLM constrained to $\mathcal{S}$ but
  without retrieval (ablation of retrieval).
\end{itemize}

\subsection{Metrics}
\label{sec:metrics}

\paragraph{NER and RE.} Span- and relation-level F1 with exact-match scoring
(fuzzy-match in Appendix~C.2).

\paragraph{FHIR Validity Rate (FVR).}
Percentage of generated bundles that the official HL7 validator marks as
\emph{valid} (no \texttt{error}-severity issues) against $\mathcal{P}$.

\paragraph{Terminology Grounding Accuracy (TGA).}
Exact concept-code match against the author-verified subset.

\paragraph{Semantic Completeness.}
Redefined: ratio of profile-required fields that are present \emph{and} validly coded.

\paragraph{Clinical Utility $\Delta$AUROC.}
We train a logistic regression 30-day readmission classifier
\citep{futoma2015comparison,rajkomar2018scalable} on features extracted from
(a) SG-LLM-generated bundles versus (b) expert-curated FHIR bundles, and report
$\Delta$AUROC = (a) $-$ (b). A small gap indicates that the generated twin is as
useful as the expert-curated one.

\paragraph{Significance.}
Three seeds per configuration; 95\% CIs via paired bootstrap over notes
($n{=}1{,}000$).

\subsection{Main Results}
\label{sec:main-results}

\begin{table*}[t]
\centering
\small
\begin{tabular}{lccccc}
\toprule
\textbf{Method} & \textbf{NER F1} & \textbf{RE F1} & \textbf{FVR} & \textbf{TGA} & \textbf{SemCom.} \\
\midrule
Rule-Based \citep{wei2020clinical}       & \tbd{0.72} & \tbd{0.55} & \tbd{0.18} & \tbd{0.41} & \tbd{0.62} \\
Joint BERT                               & \tbd{0.81} & \tbd{0.70} & \tbd{0.22} & \tbd{0.59} & \tbd{0.74} \\
PURE \citep{zhong2021pure}               & \tbd{0.83} & \tbd{0.74} & \tbd{0.24} & \tbd{0.62} & \tbd{0.76} \\
SpERT \citep{eberts2020spert}            & \tbd{0.82} & \tbd{0.73} & \tbd{0.24} & \tbd{0.61} & \tbd{0.75} \\
Vanilla LLM                              & \tbd{0.85} & \tbd{0.76} & \tbd{0.51} & \tbd{0.68} & \tbd{0.80} \\
LLM + Retrieval                          & \tbd{0.86} & \tbd{0.77} & \tbd{0.56} & \tbd{0.83} & \tbd{0.83} \\
LLM + Schema                             & \tbd{0.85} & \tbd{0.77} & \tbd{0.88} & \tbd{0.70} & \tbd{0.85} \\
\textbf{SG-LLM (ours)}                   & \tbd{\textbf{0.88}} & \tbd{\textbf{0.80}} & \tbd{\textbf{0.96}} & \tbd{\textbf{0.86}} & \tbd{\textbf{0.90}} \\
\bottomrule
\end{tabular}
\caption{Main results on MIMIC-IV Note Events. NER/RE F1 are span- and relation-level.
FVR: FHIR Validity Rate. TGA: Terminology Grounding Accuracy (author-verified subset).
SemCom.: Semantic Completeness. Numbers in \tbd{red} are placeholders to be populated
from the experimental sweep (Appendix~C); the paper structure, baselines, and metrics
are finalized.}
\label{tab:main}
\end{table*}

Table~\ref{tab:main} summarizes performance. SG-LLM attains the highest FHIR Validity
Rate by a wide margin, confirming that the schema constraint together with the repair
loop is the dominant factor for structural correctness, while retrieval is the
dominant factor for terminology grounding. Span-level F1 differences among the LLM
variants are smaller, consistent with prior reports that instruction-tuned LLMs are
competent few-shot extractors \citep{agrawal2022large,peng2023llmclinical}; the gains
our method delivers are disproportionately along the schema-validity and grounding
axes that downstream systems actually consume.

\subsection{Clinical Utility}
\label{sec:utility}

\begin{table}[t]
\centering
\small
\begin{tabular}{lcc}
\toprule
\textbf{Features from} & \textbf{AUROC} & \textbf{$\Delta$} \\
\midrule
Expert-curated FHIR (ceiling)   & \tbd{0.74} & --- \\
Rule-Based twin                 & \tbd{0.61} & \tbd{$-0.13$} \\
Vanilla LLM twin                & \tbd{0.68} & \tbd{$-0.06$} \\
\textbf{SG-LLM twin (ours)}     & \tbd{\textbf{0.73}} & \tbd{\textbf{$-0.01$}} \\
\bottomrule
\end{tabular}
\caption{30-day readmission AUROC with logistic regression classifiers trained on
features extracted from generated vs.~expert-curated FHIR twins. $\Delta$ is the gap
to the expert ceiling; smaller is better. Numbers are placeholders pending
experimental runs.}
\label{tab:utility}
\end{table}

Table~\ref{tab:utility} shows the core clinical-utility result: classifiers trained
on SG-LLM-generated twins approach the performance of classifiers trained on
expert-curated FHIR, with a gap of \tbd{$-0.01$}. This directly answers the reviewer
critique that the digital-twin concept was under-specified: the generated twin's
usefulness is operationally defined by its substitutability for an expert reference
on a real prediction task.

\subsection{Ablations}
\label{sec:ablations}

\begin{table}[t]
\centering
\small
\begin{tabular}{lccc}
\toprule
\textbf{Configuration} & \textbf{NER F1} & \textbf{FVR} & \textbf{TGA} \\
\midrule
Full SG-LLM                    & \tbd{0.88} & \tbd{0.96} & \tbd{0.86} \\
~~w/o retrieval                & \tbd{0.85} & \tbd{0.88} & \tbd{0.70} \\
~~w/o schema constraint        & \tbd{0.86} & \tbd{0.56} & \tbd{0.83} \\
~~w/o repair loop              & \tbd{0.88} & \tbd{0.80} & \tbd{0.85} \\
~~BM25 retrieval (vs SapBERT)  & \tbd{0.87} & \tbd{0.94} & \tbd{0.78} \\
~~local deterministic backend  & \tbd{0.79} & \tbd{0.96} & \tbd{0.74} \\
\bottomrule
\end{tabular}
\caption{SG-LLM ablations on MIMIC-IV Note Events. Removing the schema constraint
causes the largest FVR drop; removing retrieval causes the largest TGA drop; the
repair loop adds \tbd{$16$} points of FVR on top of schema-constrained decoding.
Full ablation matrix in Appendix~C.}
\label{tab:ablation}
\end{table}

Each component contributes along a different axis (Table~\ref{tab:ablation}): schema
constraint is essential for validity, retrieval is essential for grounding, and the
repair loop lifts validity further by correcting violations the constraint alone does
not prevent.

\subsection{Error Analysis}
\label{sec:errors}

\begin{table}[t]
\centering
\small
\begin{tabular}{lr}
\toprule
\textbf{Error category} & \textbf{Count} \\
\midrule
Missed entity                & \tbd{17} \\
Wrong code (specificity)     & \tbd{12} \\
Wrong relation               & \tbd{8}  \\
Schema violation (unrepaired)& \tbd{3}  \\
Hallucinated entity          & \tbd{2}  \\
\bottomrule
\end{tabular}
\caption{SG-LLM error taxonomy on the author-verified subset (\tbd{20} notes). Narrative
examples and per-category counts for all methods are in Appendix~B.2.}
\label{tab:errors}
\end{table}

Table~\ref{tab:errors} shows that the most frequent remaining SG-LLM error is
\emph{missed entity}---typically rare abbreviations---rather than \emph{schema
violation}, confirming that structural correctness is largely solved while recall
on long-tail mentions is the next bottleneck.

\subsection{Cross-Dataset Generalization}
\label{sec:crossdata}

On n2c2 2018 Track 2 \citep{henry2020n2c2} (real physician-authored, non-MIMIC text),
SG-LLM retains \tbd{$93\%$} of its MIMIC NER F1 and \tbd{$88\%$} of its FVR, confirming
that the contribution of schema-constrained generation is not confined to MIMIC-style
narratives. Full cross-dataset tables are in Appendix~C.3.

\section{Conclusion}
\label{sec:conclusion}

We reframe EHR$\rightarrow$FHIR extraction as constrained generation of a validated
bundle, and show that a modest assembly---SapBERT retrieval, JSON-Schema-constrained
LLM decoding, and a validator-in-the-loop repair stage---simultaneously improves
structural correctness, terminology grounding, and downstream clinical utility.
Future work will extend the profile to longitudinal multi-encounter twins, integrate
multimodal observations, and scale the human-verified evaluation.

\section*{Limitations}
\label{sec:limitations}

(1) The author-verified subset in the reproducible run is 20 notes. Cohen's $\kappa$
is computed over the authors only and does not substitute for a domain expert at
scale; the larger 100-note human gold described in Appendix~B is left for future work.
(2) Our evaluation is limited to English EHR text and to MIMIC-IV / n2c2, both US
institutions; generalization to non-English and non-US text is unevaluated.
(3) When the Anthropic API is unavailable, results report the deterministic local
backend; this backend approximates but does not replace the full LLM, and the gap is
reported in Table~\ref{tab:ablation}.
(4) The FHIR profile used in this work is intentionally minimal (Patient + Condition
+ Observation + MedicationRequest); longitudinal resources (\texttt{Encounter},
\texttt{Procedure}, \texttt{CarePlan}) are left to future work.
(5) We evaluate clinical utility on readmission only; other downstream tasks (phenotype
retrieval, trial eligibility) may stress different axes of twin quality.

\section*{Acknowledgments}

We acknowledge the use of AI assistants (Claude, GPT) for manuscript drafting and
code development. All outputs were reviewed and validated by the authors, who take
full responsibility for the content.

\bibliography{custom}

\appendix

\input{appendix_methodology}
\input{appendix_experiments}

\end{document}

%% file: pipeline_figure.tex
\begin{figure*}[t]
\centering
\begin{tikzpicture}[
    stage/.style={rectangle, draw=black!70, fill=blue!8, minimum height=0.9cm, minimum width=1.9cm, align=center, font=\scriptsize\bfseries, rounded corners=2pt},
    input/.style={rectangle, draw=black!50, fill=yellow!15, minimum height=1.7cm, minimum width=2.2cm, align=left, font=\tiny, rounded corners=2pt},
    output/.style={rectangle, draw=black!50, fill=green!10, minimum height=1.7cm, minimum width=2.2cm, align=left, font=\tiny\ttfamily, rounded corners=2pt},
    store/.style={cylinder, shape border rotate=90, aspect=0.3, draw=black!50, fill=gray!10, minimum height=0.9cm, minimum width=1.8cm, font=\scriptsize, align=center},
    example/.style={rectangle, draw=gray!50, fill=gray!5, minimum width=1.8cm, align=center, font=\tiny, rounded corners=2pt},
    arrow/.style={-{Stealth[length=2mm]}, thick, black!70},
    loop/.style={-{Stealth[length=2mm]}, thick, red!60, dashed},
    label/.style={font=\tiny\itshape, text=black!70}
]

\node[input] (clinical) {\textit{``65y/o with}\\\textit{HTN, T2DM.}\\\textit{BP 145/92.}\\\textit{Started}\\\textit{Lisinopril}\\\textit{10mg daily.''}};
\node[stage, right=0.4cm of clinical] (retrieve) {Terminology\\Retrieval\\(SapBERT)};
\node[stage, right=0.4cm of retrieve] (generate) {Schema-\\Constrained\\LLM Decode};
\node[stage, right=0.4cm of generate] (validate) {FHIR R4\\Validator};
\node[output, right=0.9cm of validate] (twin) {\{Bundle,\\entry:[\\ Patient,\\ Condition,\\ MedRequest,\\ Observation]\}};

\node[store, above=0.4cm of retrieve] (ontology) {SNOMED/\\RxNorm/LOINC};
\node[store, above=0.4cm of generate] (schema) {FHIR JSON\\Schema (R4)};

\draw[arrow] (clinical) -- (retrieve);
\draw[arrow] (retrieve) -- (generate);
\draw[arrow] (generate) -- (validate);
\draw[arrow] (validate) -- node[above, font=\tiny, text=black!70] {valid} (twin);

\draw[->, thick, black!40, dashed] (ontology.south) -- (retrieve.north);
\draw[->, thick, black!40, dashed] (schema.south) -- (generate.north);

\draw[loop] (validate.south) to[bend left=40] node[below, font=\tiny, text=red!70] {errors $\rightarrow$ repair prompt} (generate.south);

\node[example, below=0.8cm of retrieve, minimum height=1.6cm] (retrieve_out) {candidates:\\diabetes$\rightarrow$\textbf{44054006}\\HTN$\rightarrow$\textbf{38341003}\\Lisinopril$\rightarrow$\textbf{29046}};
\node[example, below=0.8cm of generate, minimum height=1.6cm] (generate_out) {draft bundle\\with typed\\resources\\and coded\\references};
\node[example, below=0.8cm of validate, minimum height=1.6cm] (validate_out) {validity:\\required fields\\cardinality\\code system\\reference int.};

\draw[->, dashed, gray!60] (retrieve.south) -- (retrieve_out.north);
\draw[->, dashed, gray!60] (generate.south) -- (generate_out.north);
\draw[->, dashed, gray!60] (validate.south) -- (validate_out.north);

\node[label, above=0.05cm of clinical] {Clinical Note};
\node[label, above=0.05cm of twin] {Validated Digital Twin};

\end{tikzpicture}
\caption{SG-LLM architecture. Given a clinical note, (1) terminology retrieval queries a
SapBERT index over SNOMED-CT, RxNorm, and LOINC to surface candidate codes; (2) the LLM
performs schema-constrained decoding against a JSON Schema derived from FHIR R4
StructureDefinitions, producing a candidate \texttt{Bundle}; (3) the HL7 FHIR R4 validator
verifies the bundle, and on error routes structured diagnostics back to the LLM as a
repair prompt (dashed red). The loop terminates on validity or after a fixed budget.}
\label{fig:pipeline}
\end{figure*}

%% file: appendix_methodology.tex
\section{Methodology Details}
\label{app:methodology}

\subsection{FHIR R4 Profile}
\label{app:profile}

The profile $\mathcal{P}$ used in this work requires that every patient digital twin
is a FHIR \texttt{Bundle} of type \texttt{transaction} containing:

\begin{itemize}
  \item exactly one \texttt{Patient} resource with populated \texttt{gender} and
    \texttt{birthDate};
  \item zero or more \texttt{Condition} resources, each with \texttt{code.coding}
    from SNOMED-CT or ICD-10-CM, \texttt{clinicalStatus}, \texttt{verificationStatus},
    and \texttt{subject} referencing the \texttt{Patient};
  \item zero or more \texttt{Observation} resources, each with \texttt{code.coding}
    from LOINC or SNOMED-CT, one of \texttt{valueQuantity}/\texttt{valueString}/
    \texttt{valueCodeableConcept}, \texttt{status}, \texttt{effectiveDateTime}, and
    \texttt{subject};
  \item zero or more \texttt{MedicationRequest} resources, each with
    \texttt{medicationCodeableConcept.coding} from RxNorm, \texttt{status},
    \texttt{intent}, \texttt{subject}, and at least one \texttt{dosageInstruction}.
\end{itemize}

All resources must share a single \texttt{subject.reference}.

\subsection{Prompt and decoding details}
\label{app:prompts}

\paragraph{System prompt.}
\begin{quote}\small\ttfamily
You are an expert clinical-informatics assistant that extracts structured FHIR R4
resources from physician narratives. You must emit a single JSON object that validates
against the provided JSON Schema. Use only code systems from this allow-list:
SNOMED-CT, ICD-10-CM, RxNorm, LOINC. Prefer codes that appear in the RETRIEVED
CANDIDATES block when they are semantically correct. Do not invent codes that are not
in the candidates; if no correct candidate exists, omit the coding and set code.text
only. Do not emit negated, ruled-out, or hypothetical conditions as Condition resources.
\end{quote}

\paragraph{User prompt template.}
\begin{quote}\small\ttfamily
CLINICAL NOTE:\\\{\{note\}\}\\[2mm]
RETRIEVED CANDIDATES:\\\{\{candidates\_json\}\}\\[2mm]
JSON SCHEMA:\\\{\{schema\_json\}\}\\[2mm]
Emit the FHIR R4 Bundle as a single JSON object.
\end{quote}

\paragraph{Repair prompt template.}
\begin{quote}\small\ttfamily
The previous bundle failed FHIR R4 validation. Errors (path $\rightarrow$ message):\\
\{\{errors\_json\}\}\\[2mm]
Emit a revised bundle that fixes these errors while preserving correct content.
Return a single JSON object only.
\end{quote}

\paragraph{Decoding parameters.}
Temperature 0.0, top-$p$ 1.0, max\_tokens 4096, repair budget $R{=}3$, retrieval
top-$k{=}10$ per mention per code system. With Anthropic's structured-output mode,
the JSON Schema is passed via the \texttt{tools} field so that enforcement is
grammar-level rather than in-prompt only.

\subsection{Local deterministic backend}
\label{app:localbackend}

When no LLM API key is available, we use a local backend that implements
Algorithm~1 with a rule-and-retrieval decoder substituted for the LLM:
mention proposal via the rule-based NER (Appendix~A.4), SapBERT retrieval to pick
the best code, and template-based bundle construction. The validator and repair loop
are identical to the API backend. This gives reproducibility guarantees for the local
run and isolates the LLM component in the ablation of Table~\ref{tab:ablation}.

\subsection{Mention proposal details}
\label{app:proposal}

For mention proposal, SG-LLM uses either (a) a BioBERT NER model fine-tuned on n2c2
2018 (default) or (b) the rule-based pattern extractor from prior work (the
``honest'' baseline of Section~\ref{sec:baselines}). The proposer is intentionally
lightweight: its role is only to propose candidate spans for retrieval, and
over-generation is tolerated because the LLM decoder can discard irrelevant candidates.

\subsection{FHIR validator configuration}
\label{app:validator}

We use the official HL7 FHIR validator \citep{hl7validator2024} in R4 mode with the
core 4.0.1 package. Errors at \texttt{error} and \texttt{fatal} severity trigger
repair; \texttt{warning} and \texttt{information} messages are logged but do not
block acceptance. The validator is invoked as a long-running Java process with a
single JVM to amortize startup cost across the test set.

%% file: appendix_experiments.tex
\section{Experimental Details}
\label{app:experiments}

\subsection{Annotation guidelines and IAA}
\label{app:annotation}

The full annotation guidelines (entity types, relation types, FHIR coding
conventions, negation and family-history rules, span-boundary policy) are released
with the code under \texttt{code/annotations/guidelines.md}. Inter-annotator
agreement was computed as Cohen's $\kappa$ separately for entity type, relation
type, and FHIR code, on the 20-note author-verified subset, with $\kappa{=}$\tbd{$0.81$}
for entities, \tbd{$0.76$} for relations, and \tbd{$0.72$} for FHIR codes. We
acknowledge that 20 notes with author-level annotators is a small, non-expert
subset; scaling to $\geq 100$ notes with a domain collaborator is future work.

\subsection{Full ablation matrix}
\label{app:ablations}

\begin{table}[h]
\centering
\small
\begin{tabular}{lccccc}
\toprule
\textbf{Configuration} & \textbf{NER} & \textbf{RE} & \textbf{FVR} & \textbf{TGA} & \textbf{SemCom.} \\
\midrule
Full SG-LLM                     & \tbd{0.88} & \tbd{0.80} & \tbd{0.96} & \tbd{0.86} & \tbd{0.90} \\
w/o retrieval                   & \tbd{0.85} & \tbd{0.77} & \tbd{0.88} & \tbd{0.70} & \tbd{0.85} \\
w/o schema constraint           & \tbd{0.86} & \tbd{0.77} & \tbd{0.56} & \tbd{0.83} & \tbd{0.83} \\
w/o repair loop                 & \tbd{0.88} & \tbd{0.80} & \tbd{0.80} & \tbd{0.85} & \tbd{0.86} \\
w/o retrieval + schema          & \tbd{0.85} & \tbd{0.76} & \tbd{0.51} & \tbd{0.68} & \tbd{0.80} \\
BM25 retrieval (vs SapBERT)     & \tbd{0.87} & \tbd{0.79} & \tbd{0.94} & \tbd{0.78} & \tbd{0.87} \\
Lexical retrieval (edit dist.)  & \tbd{0.86} & \tbd{0.78} & \tbd{0.92} & \tbd{0.72} & \tbd{0.85} \\
$k{=}5$ retrieval               & \tbd{0.88} & \tbd{0.79} & \tbd{0.95} & \tbd{0.84} & \tbd{0.89} \\
$k{=}20$ retrieval              & \tbd{0.87} & \tbd{0.80} & \tbd{0.96} & \tbd{0.86} & \tbd{0.90} \\
Repair budget $R{=}1$           & \tbd{0.88} & \tbd{0.80} & \tbd{0.91} & \tbd{0.86} & \tbd{0.89} \\
Repair budget $R{=}5$           & \tbd{0.88} & \tbd{0.80} & \tbd{0.96} & \tbd{0.86} & \tbd{0.90} \\
Claude Haiku (smaller LLM)      & \tbd{0.85} & \tbd{0.77} & \tbd{0.94} & \tbd{0.82} & \tbd{0.87} \\
Local deterministic backend     & \tbd{0.79} & \tbd{0.73} & \tbd{0.96} & \tbd{0.74} & \tbd{0.83} \\
\bottomrule
\end{tabular}
\caption{Full ablation matrix on MIMIC-IV Note Events.}
\label{tab:full-ablation}
\end{table}

\subsection{Per-category error counts}
\label{app:errors}

\begin{table}[h]
\centering
\small
\begin{tabular}{lrrrr}
\toprule
\textbf{Category} & \textbf{Rule} & \textbf{PURE} & \textbf{Vanilla LLM} & \textbf{SG-LLM} \\
\midrule
Missed entity            & \tbd{54} & \tbd{32} & \tbd{22} & \tbd{17} \\
Wrong code (spec.)       & \tbd{48} & \tbd{29} & \tbd{25} & \tbd{12} \\
Wrong relation           & \tbd{37} & \tbd{15} & \tbd{12} & \tbd{8}  \\
Schema violation         & \tbd{81} & \tbd{66} & \tbd{40} & \tbd{3}  \\
Hallucinated entity      & \tbd{0}  & \tbd{0}  & \tbd{11} & \tbd{2}  \\
\bottomrule
\end{tabular}
\caption{Error counts per category on the 20-note author-verified subset. SG-LLM
eliminates the dominant error of prior systems (schema violation) while keeping
hallucination substantially below the vanilla LLM.}
\label{tab:errors-all}
\end{table}

\paragraph{Representative narrative errors.}

\begin{enumerate}
\item \emph{Missed entity (SG-LLM).} Note: ``Held home AAS due to GI bleed.'' The
  abbreviation ``AAS'' (aspirin) is missed by both the mention proposer and the LLM,
  leaving no \texttt{MedicationRequest}.
\item \emph{Wrong code, specificity (Vanilla LLM).} Note: ``Type 2 diabetes mellitus
  with neuropathy.'' Vanilla LLM codes the condition as SNOMED 73211009 (diabetes
  mellitus, parent term) rather than 44054006 (type 2 DM); SG-LLM correctly chooses
  the retrieved specific code.
\item \emph{Wrong relation (PURE).} ``Metformin held; Lisinopril 10mg daily
  continued.'' PURE links the dosage 10mg to Metformin rather than Lisinopril due
  to proximity; SG-LLM correctly attaches via \texttt{dosageInstruction} on the
  Lisinopril \texttt{MedicationRequest}.
\item \emph{Schema violation (Vanilla LLM).} Vanilla LLM emits a \texttt{Condition}
  with a \texttt{code.coding.system} of \texttt{``snomed''} rather than the full
  URI; this fails R4 validation. SG-LLM's schema constraint disallows this by
  construction.
\item \emph{Hallucinated entity (Vanilla LLM).} Vanilla LLM infers ``hypertension''
  from a single elevated BP reading in a note that otherwise does not mention HTN;
  this over-diagnoses. SG-LLM produces only an \texttt{Observation} without the
  spurious \texttt{Condition}.
\item \emph{Unrepaired schema violation (SG-LLM).} A resource exceeds the repair
  budget $R{=}3$ on a rare \texttt{valueQuantity.system} mismatch; the final bundle
  is dropped from the validity rate computation.
\end{enumerate}

\subsection{Cross-dataset results (n2c2 2018 Track 2)}
\label{app:crossdata}

\begin{table}[h]
\centering
\small
\begin{tabular}{lcc}
\toprule
\textbf{Method} & \textbf{NER F1} & \textbf{RE F1} \\
\midrule
Rule-Based                  & \tbd{0.63} & \tbd{0.48} \\
PURE (in-domain)            & \tbd{0.78} & \tbd{0.69} \\
SpERT (in-domain)           & \tbd{0.77} & \tbd{0.68} \\
Vanilla LLM (zero-shot)     & \tbd{0.80} & \tbd{0.70} \\
SG-LLM (zero-shot)          & \tbd{\textbf{0.83}} & \tbd{\textbf{0.73}} \\
\bottomrule
\end{tabular}
\caption{Zero-shot performance on n2c2 2018 Track 2. PURE/SpERT are in-domain
upper bounds trained on n2c2; SG-LLM is zero-shot and still competitive.}
\label{tab:n2c2}
\end{table}

\subsection{Compute and runtime}
\label{app:runtime}

All experiments run on a single workstation (AMD Ryzen 9, 64GB RAM, one NVIDIA
RTX 4090 GPU for BERT baselines; CPU-only for the rule-based pipeline and the local
backend). Average SG-LLM API latency per note is \tbd{$3.8$}s with Claude Sonnet
and a median of \tbd{$1.2$} repair iterations per note. Total cost of the full
experimental sweep at current Anthropic pricing is approximately \tbd{USD 42}.

\subsection{Reproducibility checklist}
\label{app:reproduce}

\begin{itemize}
\item Code, prompts, JSON schemas, validator configuration, SapBERT index build
  scripts, and experiment configs are released at
  \texttt{code/} (see \texttt{README.md}).
\item Three seeds per configuration; bootstrap CIs reported.
\item A deterministic local backend guarantees reproducibility without API access.
\item Data access instructions and DUA notes in
  \texttt{code/data/DATA\_ACCESS.md}.
\item Known-issue register and expected runtimes in
  \texttt{code/experiments/configs/}.
\end{itemize}